\documentclass[11pt,a4paper]{article}
\usepackage{times,latexsym}
\usepackage{url}
\usepackage[T1]{fontenc}

\usepackage[acceptedWithA]{tacl2021v1}
\usepackage{xspace,mfirstuc,tabulary}

\newif\iftaclinstructions
\taclinstructionsfalse 
\iftaclinstructions
\renewcommand{\confidential}{}
\renewcommand{\anonsubtext}{(No author info supplied here, for consistency with
TACL-submission anonymization requirements)}
\newcommand{\instr}
\fi

\iftaclpubformat 

\else

\fi

\usepackage{array}
\usepackage{times}
\usepackage{latexsym}
\usepackage[T1]{fontenc}
\usepackage[utf8]{inputenc}
\usepackage{graphicx}
\usepackage{microtype}
\usepackage{inconsolata}
\usepackage{hyperref}
\usepackage{subcaption}
\usepackage{bbding}
\usepackage{pifont}
\usepackage{amsmath}
\usepackage{multirow}
\usepackage{multirow, booktabs, multicol}
\usepackage{xspace}
\usepackage[normalem]{ulem}
\usepackage{listings}

\definecolor{lightgray}{gray}{0.9}
\definecolor{darkgray}{gray}{0.4}
\definecolor{purple}{rgb}{0.58,0,0.82}

\lstdefinelanguage{plan}{
  sensitive=true,
  morecomment=[l]{//},
  morestring=[b]"
}

\usepackage{comment}
\newcommand{\wiki}{\textit{WikiEvents}\xspace}
\newcommand{\muc}{\textit{MUC}\xspace}
\newcommand{\cmnee}{\textit{CMNEE}\xspace}
\newcommand{\triggers}{\textit{Triggers}\xspace}
\newcommand{\trigger}{\textit{trigger}\xspace}
\newcommand{\triggerplural}{\textit{triggers}}
\newcommand{\tanlsys}{\textsc{TANL}\xspace}
\newcommand{\tanl}{\textsc{TANL}$^\diamondsuit$\xspace}
\newcommand{\gttsys}{\textsc{GTT}\xspace}

\newcommand{\uiucsys}{\textsc{GenIE}\xspace}

\newcommand{\degreesys}{\textsc{Degree}\xspace}

\newcommand{\gptomin}{\textsc{GPT-4o-mini}\xspace}
\newcommand{\gpto}{\textsc{GPT-4o}\xspace}

\newcommand{\human}{\textsc{Human}\xspace}
\newcommand{\random}{\textsc{Random}\xspace}

\colorlet{NextBlue}{red!222!green!48!blue!99}

\title{Are \textit{Triggers} Needed for Document-Level Event Extraction?}

\author{
  Shaden Shaar\thanks{These authors contributed equally to this work}, 
  Wayne Chen\footnotemark[1], 
  Maitreyi Chatterjee, 
  \textbf{ Barry Wang}, \\
  \textbf{Wenting Zhao and }
  \textbf{Claire Cardie}
  \\
  \ \\
  Cornell University, USA
  \\
  \texttt{\{ss2753, zc272\}@cornell.edu}
}

\date{}

\begin{document}
\maketitle

\begin{abstract}

Most existing work on event extraction has focused on sentence-level 
texts and presumes the identification of a \trigger-span --- a word or 
phrase in the input that evokes the occurrence of an event of interest. 
Event arguments are then extracted with respect to the trigger.
Indeed, triggers are treated as integral to, and trigger detection as an essential component of, event extraction. 
In this paper, we provide the first investigation of
the role of triggers for the more difficult and much less studied task of document-level event extraction.
We analyze their usefulness in multiple end-to-end and pipelined transformer-based event extraction models for three document-level event extraction datasets, measuring performance using triggers of varying quality (human-annotated, LLM-generated, keyword-based, and random).
We find that whether or not systems benefit from explicitly extracting triggers depends both on dataset characteristics (i.e. the typical number of events per document) and task-specific information available during extraction (i.e. natural language event schemas). Perhaps surprisingly,
we also observe that the mere existence of triggers in the input, even random ones, is important for prompt-based in-context learning approaches to the task.
\end{abstract}

\section{Introduction}\label{sec:intro}

\begin{figure*}[tbh]
    \centering
    \includegraphics[width=\textwidth]{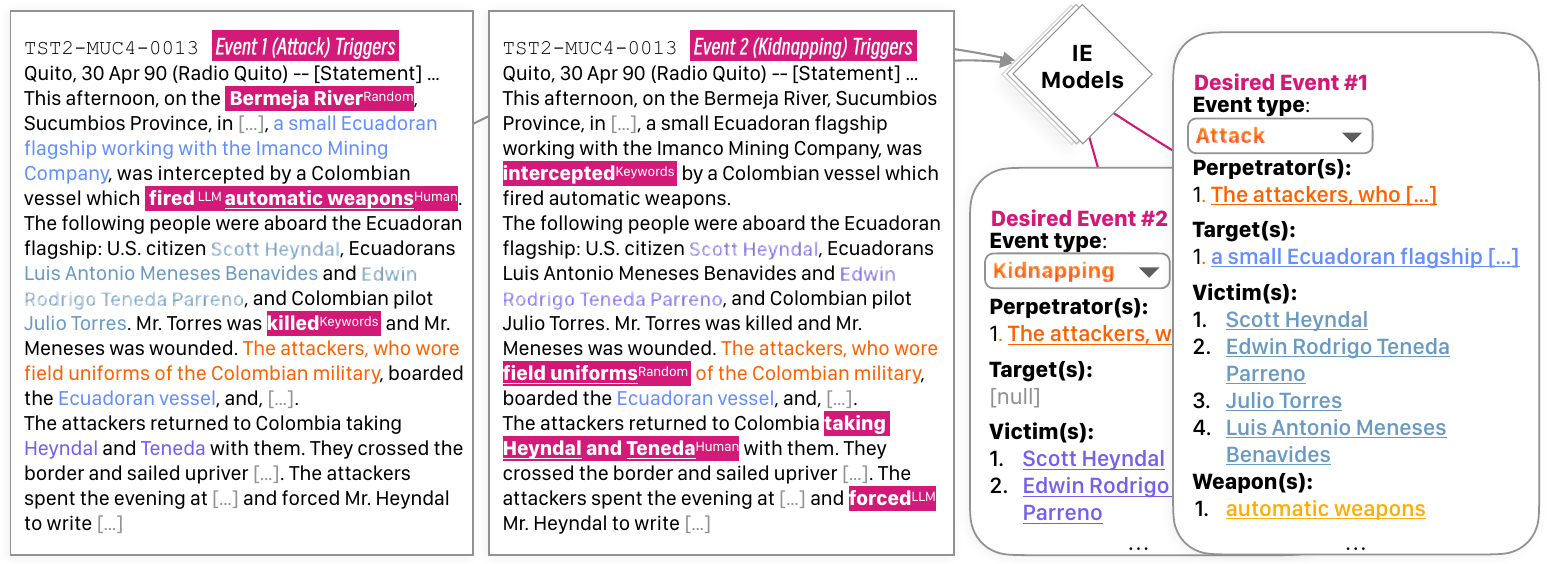}
    \vspace{-18pt} 
    \caption{\textbf{Document-level event extraction example from \muc.} {\small Two events 
    should be extracted from the given text, an \textsc{Attack} (doc on left) and a \textsc{Kidnapping} (doc on right). Note that only one mention (among many coreferent ones) per event argument is shown. 
    Some event arguments are empty and some might have multiple fillers. The (duplicated) document depicts event triggers obtained from different annotation sources (labeled with pink background $^{\text{and the annotation source}}$); more details in Section~\ref{sec:approach}.}
    }
    \label{fig:overview}
\end{figure*}

There is a long history of research in Natural Language Processing (NLP) 
on the topic of \textit{event extraction}~\citep{DBLP:journals/nle/Grishman19}. The 
vast majority of that research focuses on \textit{sentence-level event 
extraction}~\citep[][i.a.]{ji-grishman-2008-refining, li-etal-2013-joint, yang-etal-2019-exploring, Wadden2019EntityRA}: 
given a text and a predefined set of \textit{event type}s $E$, each $e \in E$ with 
its own set of \textit{arguments} $A_e$  (also known as \textit{roles}), find all
role-fillers of these events in the text, identifying for each (1) the 
\textit{trigger} span $t$ that denotes the occurrence of the event, 
(2) the event type $e$, and (3) a span of text \textit{s}
for each instantiated event argument $a \in A_e$ (also called \textit{role fillers}) mentioned \textbf{within the sentence containing} \textit{t}.
Prior work treats triggers as integral to accurate event 
extraction~\citep{mhamdi-etal-2019-contextualized,tong-etal-2020-improving,lin-etal-2022-trend}. 
Specifically, trigger detection is typically the first step of event extraction
as it informs event type categorization and provides a lexical anchor for argument
extraction.  Thus, trigger spans are marked during the data annotation process for sentence-level event extraction~\citep{doddington2004automatic}.

In recent years, however, there has been increasing interest in \textit{document-level event 
extraction}: given a document $d$, a set of event types $E$, and their associated arguments 
$A_{e \in E}$, produce one filled event template for each relevant event described in $d$ with event 
arguments recognized at the \textit{entity level}, i.e., via a single representative
mention.\footnote{Other names for
the task include template filling \cite{jurafskyspeech}, generalized template extraction 
\citep{chen-etal-2023-iterative}, N-ary relation extraction \cite{jain-etal-2020-scirex}.}
%
Document-level event extraction is important because the vast majority of 
real-world news and narrative texts describe multiple complex interacting events.
\citet{li-etal-2021-document}, for example, find that the arguments of close to
40\% of events in their Wikipedia-based dataset appear in sentences other than
the event trigger sentence.
Unfortunately, the document-level task setting poses significant challenges \citep{das-etal-2022-automatic}. Documents can contain zero events or
discuss multiple events of the same or similar types. Events and event arguments 
are likely to be mentioned many times throughout the document, requiring coreference resolution. Arguments can be shared across events.  Information about different events can be intermingled. 


For example, in Figure~\ref{fig:overview}, ``Heyndal'', and 
``Teneda'' are arguments for both the \textsc{Attack} and \textsc{Kidnapping} 
events; and their multiple mentions in the text
must be recognized as coreferent so that only one mention
for each is included in each of the output events. Additionally, the arguments for each 
event are spread across multiple sentences.

Note that event trigger-span identification is not required as part of the output; indeed,
most datasets for the task do not include trigger-span 
annotations~\cite{better, jain-etal-2020-scirex, pavlick-etal-2016-gun}. 
On one hand, this makes sense: a single event may have multiple triggers (or possibly none), and conversely a particular trigger text span may be associated with more than one event. Also, it is known that event individuation --- 
the problem of distinguishing distinct events --- is difficult even for human experts~\cite{gantt-etal-2023-event}.
On the other hand, event triggers, which are by definition lexical indicators of an
event, should help with event type identification and systems may benefit from extracting them as an intermediate step.  In cases where they denote a
(verb or noun) predicate for the event, they will serve as an anchor to localize 
argument extraction~\cite{ebner-etal-2020-multi}. Finally, consecutive triggers denoting
different event types will be cues to segment the text into event-specific chunks.

In this paper, to the best of our knowledge, we provide the first investigation of the usefulness of triggers for document-level event extraction. 
As described in Section~\ref{sec:approach}, we treat \triggers as ``rationales" that indicate
the presence of an event and guide the system in finding
the argument of the event.
We analyze whether using triggers improves event extraction performance in multiple end-to-end and pipelined transformer-based \cite{NIPS2017_3f5ee243} event extraction models for three document-level event extraction datasets. In particular, we measure event extraction performance for a range of event triggers of varying quality (human-annotated, LLM-generated, keyword-based, and random). 
Additionally, we compare the results to in-context learning prompting baselines. 
We publicly release all trigger annotations used in our experiments,\footnote{\hyperlink{https://github.com/WayneChen2021/Triggers-Data.git}{GitHub Link}} including our manually annotated trigger-spans for the classic \muc dataset.

We find that the usefulness of triggers depends, in part, on characteristics of the extraction task and
its data, i.e., the density of relevant events in each document and the degree to which event arguments are localized with respect to the trigger.  We also find that lower-quality, automatically identified triggers can be used as an alternative to human annotations. This is good news given the cost of obtaining manual annotations.  In addition, we show that performance robustness in the face of degraded trigger quality can be achieved by
making descriptive event information available to the extraction model.  Perhaps surprisingly,
we also find that the mere availability of triggers, even random ones, is important for prompt-based in-context learning approaches to the task. See Section \ref{subsec:trigger-creation} for access to our public data and code. 

\begin{figure*}[tbh]
    \centering
    \small
    \includegraphics[width=1.05\textwidth]{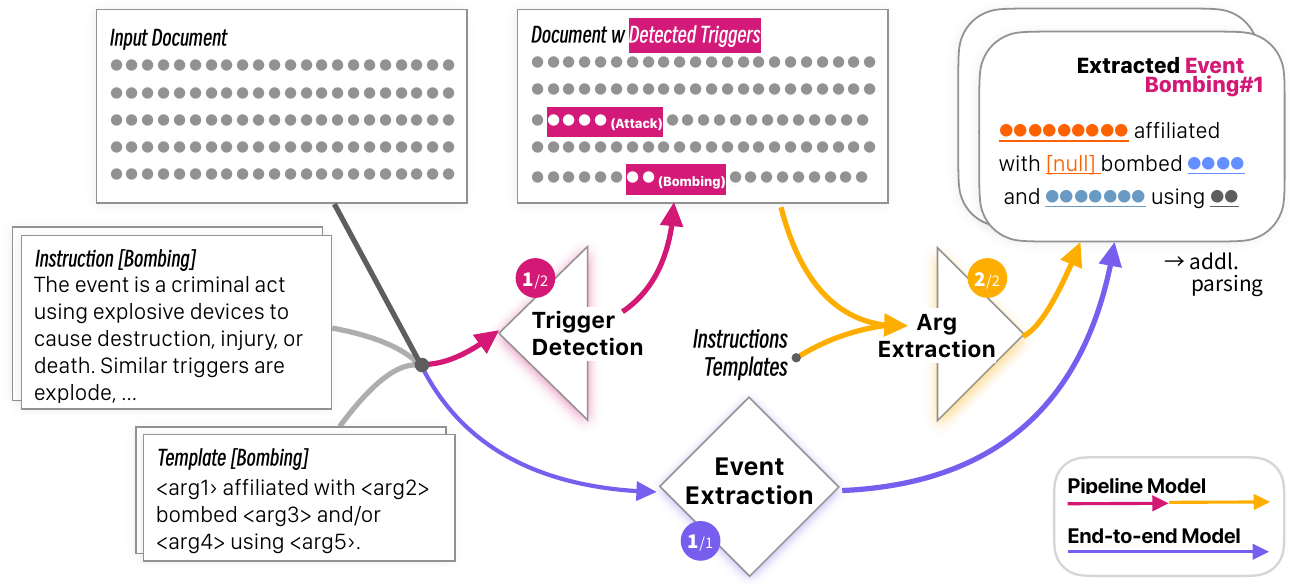}
    \caption{{\small \textbf{Approach Overview}:  We examine four state-of-the art event extraction systems (e.g., \tanlsys, \degreesys, \gttsys, \uiucsys). 
    We run these models in two
    architectures, when possible, \textit{pipeline} and end-to-end (\textit{E2E}). We also 
    include comparisons to $\textsc{GPT-4o}$ and $\textsc{GPT-4o-mini}$ prompted via in-context few-shot 
    learning (not shown). Models differ in the task-specific information they provide---document only (\tanlsys, \gttsys), document+instructions (\tanlsys+\uiucsys), 
    document+instructions+template (\degreesys).}
    } 
    \label{fig:models-arch}
\end{figure*}

\section{Related Work}\label{sec:related-work}


\paragraph{Document-level Template Extraction.} Doc\-u\-ment-lev\-el event extraction was introduced in the Message Understanding Conferences (\muc) 
\cite{sundheim-1991-overview} as a task to identify occurrences of predefined 
event types and fill the associated template roles by extracting entity mentions from the text or 
categorizing aspects of the event. Other event extraction datasets include \textit{RAMS} \cite{ebner-etal-2020-multi}, \cmnee \cite{zhu-etal-2024-cmnee}, \wiki \cite{li-etal-2021-document}, and \textit{MAVEN} \cite{wang-etal-2020-maven}. We focus our analysis on \muc, \wiki, and \cmnee due to their long-document, multi-event nature and because their limited selection of event types enables clearer analysis of performance differences between our benchmarked systems. Document-level event extraction systems utilize drastically different techniques: earlier methods relied on pattern matching \cite{autoslog, multi_level_bootstrap}, while more modern approaches use deep learning to represent event components such as triggers and entities \cite{Wadden2019EntityRA, wang-etal-2023-document, xu-etal-2021-document}. Most recently, the task has been framed as sequence-to-sequence generation (i.e. \tanlsys \cite{tanl}, \gttsys \cite{du-etal-2021-template} and \degreesys \cite{hsu-etal-2022-degree}) built upon powerful transformer \cite{NIPS2017_3f5ee243} language models. Our work focuses primarily on such methods due to the flexibility of sequence-to-sequence frameworks \cite{tanl} and because we believe future methods will increasingly utilize constantly improving LLMs.

\vspace{-1ex}

\paragraph{Event Argument Extraction.} 
Argument extraction is a subtask of event extraction that requires identifying
fillers for the roles associated with the event given an event type as well as its \trigger span \cite{li-etal-2021-document, ren-etal-2023-retrieve, ma-etal-2022-prompt}. Argument extractors are commonly employed in event extraction systems that segment the task into trigger/event detection and typing followed by argument extraction. We primarily focus on \tanlsys \cite{tanl} and \uiucsys \cite{li-etal-2021-document} that model the task via sequence generation, though other methods that rely on careful manipulation and learning of embeddings also exist \cite{chen-etal-2015-event, yang-etal-2019-exploring}.


\section{Approach}\label{sec:approach}

Many NLP systems predict \textit{rationales} 
in conjunction with outputs as a
means of explaining them.
In this study, we effectively treat \triggerplural~as rationales: a system is given, or predicts,
event triggers that may guide identification of associated arguments. 
Previous studies have illustrated that NLP systems can leverage rationales in various ways. 
\citet{deyoung-etal-2020-eraser} assert that rationales can function as explanations, helping
end users evaluate a system's trustworthiness. \citet{yang2018hotpotqa} show that providing 
rationales as extra supervision enhances a model's capability to perform multi-hop reasoning. 
Most recently, \citet{chen-etal-2022-rationalization} demonstrate the potential of 
rationale-based NLP systems on adversarial inputs. 
Given these benefits, we investigate whether triggers can offer similar advantages to 
document-level event extraction systems 
and prompting-based baselines. 

\triggers, as defined in \citet{ace05}, 
refer to text spans (verbs, nouns, or adjectives) that most clearly express the occurrence of one or
more events.  
Since trigger annotation is potentially a time-consuming and expensive process
and trigger quality is likely to affect their efficacy for document-level
event extraction,
\textbf{ we experiment with four sources of triggers}: human-generated, LLM-generated, keyword-based, and randomly selected.  Trigger generation and datasets will be explained in Section~\ref{sec:dataset}.

Our approach for the evaluation is depicted in Figure~\ref{fig:models-arch}.
Specifically, we study triggers in 
conjunction with \textbf{four recent sequence-to-sequence transformer-based approaches to document-level event 
extraction}: \textsc{TANL} \cite{tanl},
\textsc{GTT} \cite{du-etal-2021-template}, \textsc{Degree} \cite{hsu-etal-2022-degree}, and \textsc{GenIE} \cite{li-etal-2021-document}.  
In addition to adopting different underlying neural architectures, each has
input requirements that differ in the amount of task-specific knowledge they expect. The models and input settings are described in Section~\ref{sec:models}.
We consider both \textbf{end-to-end} and
\textbf{pipelined} system architectures (Section~\ref{sec:model-type}) and
provide comparisons to \textbf{\gpto and \gptomin baselines} prompted via in-context learning. Details on the experimental setup can be
found in Section~\ref{sec:experiments}.
Results are presented and analyzed in Section~\ref{sec:result-discussion}.


We specifically investigate the following:
\begin{enumerate}
\item Do corpora characteristics, such as the typical number of events and the distribution of event arguments across a document, influence the utility of triggers? For example, triggers might be useful anchors for argument extraction when there are many events with arguments located in close proximity to triggers.
\item Can we view triggers as grounding information for the event extraction task, and if so, could other task/corpus specific information be utilized in place of triggers such that trigger-free methods can match trigger-based methods? For example, can methods that ingest fine-grained event schema details without explicitly identifying triggers during extraction be as performant as methods explicitly learning trigger identification? If grounding information beyond triggers is effective and can be easily constructed, we may be able to bypass expensive human trigger annotation efforts during event extraction dataset creation.
\item To further minimize manual annotation, perhaps only providing in-context examples to a modern LLM can achieve reasonable performance. Alternatively, the many subtasks subsumed in event extraction (event detection, entity detection, entity and event typing) may still be too complex for LLMs. We also explore how triggers impact in-context learning methods.
\item Event extraction methods are generally either pipelined (extract a trigger, then extract a single event's arguments conditioned on a trigger) or end-to-end (the model learns trigger and argument extraction simultaneously). Which performs better? Does the source of the trigger annotations impact performance between the two paradigms? While our benchmark methods by default fit exclusively in one category, slight modifications allow single systems to model both paradigms such that we can center our analyses around the paradigm used.
\end{enumerate}
\begin{table*}[tbh!]
    \centering
    \small
    \resizebox{\textwidth}{!}{  
    \begin{tabular}{l@{ }@{ }@{ } c c c c c c c c@{}}
        \toprule
        \bf Dataset & \textbf{\# docs} & \textbf{\# tokens /} & \textbf{\# events /} & \textbf{\% docs w/} & \textbf{\# events /} & \textbf{\# args /} &  \textbf{\% filled args} & \textbf{\# tokens} \\
                    &  & \textbf{doc} & \textbf{doc} & \textbf{$\geq$ 1 event} & \textbf{relevant doc} & \textbf{event} & \textbf{/ event} & \textbf{between args} \\
        \midrule
        \multicolumn{8}{l}{\bf \muc} \\
        -- train & 1300 & 2089.0 & 0.86 & 57\% & 1.59 & 2.13 & 43\% & 33.9\\
        -- val & 200 & 2153.4 & 0.95 & 58\% & 1.65 & 2.28 & 46\% & 44.8\\
        -- test & 200 & 1979.8 & 1.02 & 63\% & 1.63 & 2.34 & 47\% & 38.3 \\
        \midrule
        \multicolumn{8}{l}{\bf \wiki}\\
        -- train & 206 & 4162.9 & 15.73 & 95\% & 16.71 & 1.26 & 28\% & 1.6\\
        -- val & 20 & 3422.2 & 17.25 & 100\% & 17.25 & 1.39 & 25\% & 1.5 \\
        -- test & 20 & 3785.0 & 18.25 & 95\% &  19.21 & 1.08 & 30\% & 1.9\\
        \midrule
        \multicolumn{8}{l}{\bf \cmnee}\\
        -- train & 6448 & 192.2 & 0.95 & 58\% & 1.65 & 5.02 & 100\% & 8.8\\
        -- val & 1162 & 193.3 & 0.97 & 66\% & 1.47 & 4.35 & 100\% & 8.4\\
        -- test & 1370 & 188.2 & 1.60 & 80\% & 2.00 & 6.12 & 100\% & 9.2\\
        \bottomrule
    \end{tabular}
    }
    
\caption{{\small \textbf{Dataset Statistics.} This shows { \textit{\# tokens / doc}} and { \textit{\# events / doc}} on average; and for documents with events, the average { \textit{\# events / relevant doc}} and { \textit{\# args / event}}. It also includes { \textit{\% filled args / event}} and the standard deviation of \# tokens between arguments and
the midpoint of the text chunk that covers all of the event's arguments
({ \textit{\# tokens between args}}).}}
    \label{table:datastat-stats}
\end{table*}

\vspace{-1ex}
\section{Datasets and Trigger Creation}\label{sec:dataset}

Similar to previous work, we perform our analyses on the \muc \cite{sundheim-1991-overview}, \wiki \cite{li-etal-2021-document} 
and \cmnee \cite{zhu-etal-2024-cmnee} datasets. 
These are described next, followed by the event trigger
generation process.

\subsection{Datasets}
\paragraph{\muc}\label{sec:muc-dataset} is a dataset introduced to study event extraction in English news articles reporting Latin American terrorist incidents. The dataset consists of document-template(s) pairs covering six event types.
A document can contain no events or multiple events; similarly, an argument role can have no argument entities or multiple entities. 
Following previous work \citep{ du-etal-2021-template, wang-etal-2023-probing}, we restricted event templates to contain 6 out of the original 24 roles --- \textsc{event-type}, \textsc{perpetrator}, \textsc{perpetrator-organization}, \textsc{target}, \textsc{victim}, \textsc{weapon}. \muc does not provide any \trigger-span annotations, but we
obtain \human (manually identified) trigger annotations
as part of this work.  The annotation process is described in Section~\ref{subsec:trigger-creation}.

\paragraph{\wiki}\label{sec:wiki-dataset} 
is a collection of English Wikipedia texts scraped from its ``event" pages. 
The dataset consists of document-template(s) pairs covering 33 event types.
In contrast to \muc, each role is filled by at most one entity.
\wiki provides \textsc{human} (manually identified) \trigger-span annotations. 

\paragraph{\cmnee}\label{sec:cmnee-dataset} 
is a Chinese dataset of reports of military activity. 
The dataset consists of document-template(s) pairs covering eight event types; argument roles always have at least one entity filler. 
Since the dataset is in Chinese, we translated the documents to English using  the \hyperlink{https://pypi.org/project/googletrans/}{Google Translate API}.
We used HTML tags to delimit entity mentions before passing it through the translator to recover their translated correspondences.
Because tags are sometimes dropped during translation, some mentions are unrecoverable.  We filter out documents with such issues resulting in retaining about ~60\% of all documents. \cmnee provides \textsc{human} \trigger-span annotations.

\subsection{Dataset Analysis}\label{sec:dataset-analysis}

The characteristics of the datasets vary considerably.  Table~\ref{table:datastat-stats} shows that \wiki has many events per document compared to \muc and \cmnee. \muc has few events on average per document (many documents have no associated events) whereas all of the \wiki documents and 62\% of the \cmnee documents contain at least one event. 

Finally, the \textit{localization} of the argument
mentions for event 
descriptions in the documents of each dataset varies
tremendously --- for \wiki, the arguments for each
event are 1-2 tokens apart\footnote{We used nltk.word\_tokenize to tokenize documents. For each event, we identify the index of the each of its arguments' first token, then compute the standard deviation of these indices.} on average;  for \cmnee, arguments for each event are generally within 8-9 tokens
of each other;  arguments for \muc events, on the other hand, are very spread out, spanning about 40 tokens.

Ideally the documents in our datasets would both be long and have event arguments widely dispersed throughout the texts, but such a dataset does not exist at the time of writing. Our selection of \muc, \wiki, and \cmnee as a group nevertheless still allows us to perform meaningful analyses of these characteristics.

\subsection{Trigger Creation}\label{subsec:trigger-creation}

We investigated four sources of trigger annotation: \textsc{human}, \textsc{LLM}, \textsc{keyword}, and \textsc{random} and provide details on how each is generated below. 
Examples of each trigger source are provided in Figure~\ref{fig:overview}. 
We will make publicly available all trigger sets employed in our experiments.\footnote{\hyperlink{GitHub Link}{GitHub Link} (not available during anonymous review)}

\paragraph{\textsc{Human} triggers} \textsc{human} trigger spans 
are obtained by hiring annotators to identify the most informative trigger for each event in the document. They are the most expensive trigger type to obtain.
\wiki and \cmnee provide one human-annotated trigger per event;  \muc does not.
As a result, we hired 9 annotators to identify all plausible trigger spans
for each document-event pair.
57\% of the document-template pairs in the dataset were annotated by at 
least two annotators. Average $\alpha$-Krippendorff and $F$-Kappa agreement scores across annotators of $0.530$ and $0.453$, respectively, indicate a moderate level of agreement. The annotation guidelines are provided in Appendix~\ref{appendix:annotation-guideline}.


To obtain a single trigger for each document-event pair, we take the union of all annotated triggers for the event and select the first, i.e.\ earliest, span among them. This typically produced better extraction scores than using the trigger span most centrally located among the event's arguments or other plausible trigger selection heuristics.

\paragraph{\textsc{LLM} triggers}
Recent work has demonstrated that \textsc{GPT} performs well on some information
extraction tasks \citep{wang-etal-2023-code4struct, wadhwa-etal-2023-revisiting}. We therefore hypothesized LLM-generated triggers might be an acceptable proxy for \textsc{Human} annotated triggers. We use zero-shot prompting of \textsc{GPT-4-turbo} to identify and rank multiple trigger-span(s) for each event and select the highest ranking span.\footnote{The prompts used will be part of
the GitHub release.}

\paragraph{\textsc{Keyword} triggers}
\textsc{Keyword} triggers are obtained by identifying a set of event-specific keywords at the corpus level and applying them heuristically for each document. 
For each event type $e \in E$, we identified documents that contained at least one event of type \textit{e} to form a list $L_{\textit{e}}$ of words associated with \textit{e}.  We manually filtered each $L_{\textit{e}}$ producing lists $\hat{L}_{\textit{e}}$ that contain only the words that are plausible triggers for \textit{e} (i.e., ``exploded" or ``bombed" for a bombing event), thus forming a list of keywords for each event type. We then use the entries from $\hat{L}_{\textit{e}}$ to annotate all events of type \textit{e}. 

\paragraph{\textsc{Random} triggers}
We also collected \random triggers by randomly selecting a span of 1-3 words from the document. They are obviously
of poor quality and are used as a control to identify whether the mere presence of triggers in the system input is
helpful.


\begin{table*}[t]
        \centering
        \resizebox{.66\textwidth}{!}{
        \begin{tabular}{@{}l@{}r*{3}r@{}}
            \toprule
            \multicolumn{1}{p{0.23\linewidth}}{\bf Experiment} & \multicolumn{1}{c}{\bf \muc} & \multicolumn{1}{c}{\bf \wiki} & \multicolumn{1}{c}{\bf \cmnee}\\
            \midrule 
            {} & \multicolumn{3}{c}{\bf no \triggers}\\
            \midrule 

$\textsc{GPT-4o-mini}$$_{prompting}$ & $.434_{\pm .01}$  & $.325_{\pm .02}$  & $.565_{\pm .00}$  & \\
$\textsc{{GPT-4o}}$$_{prompting}$ & $.391_{\pm .01}$  & $.339_{\pm .02}$  & $.635_{\pm .00}$  & \\
$\textsc{{TANL}}_{e2e}$ &\colorbox{magenta!5}{ $.559_{\pm .01}$  }& $.361_{\pm .02}$  & $.438_{\pm .00}$  & \\
$\textsc{{DEGREE}}_{e2e}$ &\colorbox{magenta!95}{ $.597_{\pm .00}$  }& $.248_{\pm .01}$  &\colorbox{magenta!95}{ $.789_{\pm .00}$  }& \\
$\textsc{{GTT}}_{e2e}$ & $.454_{\pm .01}$  & \multicolumn{1}{c}{---}  & $.318_{\pm .00}$  & \\
    
            \midrule 
            {} & \multicolumn{3}{c}{\bf Human \triggers}\\
            \midrule 
    
$\textsc{GPT-4o-mini}$$_{prompting}$ & $.438_{\pm .02}$  & $.364_{\pm .02}$  & $.564_{\pm .00}$  & \\
$\textsc{{GPT-4o}}$$_{prompting}$ & $.405_{\pm .01}$  & $.379_{\pm .02}$  & $.667_{\pm .00}$  & \\
$\textsc{{TANL}}_{pipeline}$ & $.460_{\pm .01}$  &\colorbox{magenta!95}{ $.471_{\pm .02}$  }& $.596_{\pm .01}$  & \\
$\textsc{{DEGREE}}_{pipeline}$ & $.524_{\pm .01}$  & $.357_{\pm .01}$  &\colorbox{magenta!35}{ $.723_{\pm .00}$  }& \\
$\textsc{TANL}$+$\textsc{GENIE}_{pipeline}$ & $.464_{\pm .01}$  & $.382_{\pm .01}$  & $.599_{\pm .00}$  & \\
$\textsc{{TANL}}_{e2e}$ & $.530_{\pm .01}$  & $.337_{\pm .02}$  & $.410_{\pm .00}$  & \\
$\textsc{{DEGREE}}_{e2e}$ &\colorbox{magenta!35}{ $.566_{\pm .01}$  }& $.339_{\pm .02}$  &\colorbox{magenta!65}{ $.782_{\pm .00}$  }& \\
$\textsc{{GTT}}_{e2e}$ & $.429_{\pm .01}$  & ---  & $.326_{\pm .00}$  & \\
    
            \midrule 
            {} & \multicolumn{3}{c}{\bf LLM \triggers}\\
            \midrule 
    
$\textsc{GPT-4o-mini}$$_{prompting}$ & $.416_{\pm .02}$  & $.368_{\pm .02}$  & $.580_{\pm .00}$  & \\
$\textsc{{GPT-4o}}$$_{prompting}$ & $.414_{\pm .01}$  &\colorbox{magenta!5}{ $.383_{\pm .02}$  }& $.659_{\pm .00}$  & \\
$\textsc{{TANL}}_{pipeline}$ & $.305_{\pm .01}$  & $.207_{\pm .01}$  & $.543_{\pm .00}$  & \\
$\textsc{{DEGREE}}_{pipeline}$ &\colorbox{magenta!65}{ $.567_{\pm .01}$  }& $.335_{\pm .01}$  &\colorbox{magenta!5}{ $.715_{\pm .00}$  }& \\
$\textsc{TANL}$+$\textsc{GENIE}_{pipeline}$ & $.490_{\pm .01}$  &\colorbox{magenta!35}{ $.388_{\pm .01}$  }& $.558_{\pm .00}$  & \\
$\textsc{{TANL}}_{e2e}$ & $.507_{\pm .00}$  & $.267_{\pm .02}$  & $.440_{\pm .00}$  & \\
$\textsc{{DEGREE}}_{e2e}$ & $.557_{\pm .00}$  & $.313_{\pm .02}$  & $.703_{\pm .00}$  & \\
$\textsc{{GTT}}_{e2e}$ & $.439_{\pm .02}$  & \multicolumn{1}{c}{---}  & $.332_{\pm .00}$  & \\

            \midrule 
            {} & \multicolumn{3}{c}{\bf Keyword \triggers}\\
            \midrule 
    
$\textsc{GPT-4o-mini}$$_{prompting}$ & $.429_{\pm .02}$  & $.361_{\pm .02}$  & $.582_{\pm .00}$  & \\
$\textsc{{GPT-4o}}$$_{prompting}$ & $.420_{\pm .01}$  &\colorbox{magenta!65}{ $.389_{\pm .02}$  }& $.659_{\pm .00}$  & \\
$\textsc{{TANL}}$$_{pipeline}$ & $.525_{\pm .02}$  & $.306_{\pm .02}$  & $.640_{\pm .00}$  & \\
$\textsc{{DEGREE}}$$_{pipeline}$ & $.532_{\pm .01}$  & $.241_{\pm .01}$  & $.532_{\pm .00}$  & \\
$\textsc{TANL}$+$\textsc{GENIE}_{pipeline}$ & $.470_{\pm .01}$  & $.316_{\pm .01}$  & $.628_{\pm .00}$  & \\
$\textsc{{TANL}}$$_{e2e}$ & $.495_{\pm .00}$  & $.292_{\pm .02}$  & $.443_{\pm .00}$  & \\
$\textsc{{DEGREE}}$$_{e2e}$ & $.523_{\pm .01}$  & $.273_{\pm .01}$  & $.601_{\pm .00}$  & \\
$\textsc{{GTT}}$$_{e2e}$ & $.455_{\pm .01}$  & \multicolumn{1}{c}{---}  & $.324_{\pm .00}$  & \\
    
            \midrule 
            {} & \multicolumn{3}{c}{\bf Random \triggers}\\
            \midrule 
    
$\textsc{GPT-4o-mini}$$_{prompting}$ & $.412_{\pm .01}$  & $.369_{\pm .02}$  & $.533_{\pm .01}$  & \\
$\textsc{{GPT-4o}}$$_{prompting}$ & $.402_{\pm .00}$  & $.359_{\pm .02}$  & $.533_{\pm .01}$  & \\
$\textsc{{TANL}}$$_{pipeline}$ & $.116_{\pm .01}$  & $.059_{\pm .01}$  & $.223_{\pm .00}$  & \\
$\textsc{{DEGREE}}$$_{pipeline}$ & $.538_{\pm .01}$  & $.255_{\pm .01}$  & $.589_{\pm .00}$  & \\
$\textsc{TANL}$+$\textsc{GENIE}_{pipeline}$ & $.322_{\pm .01}$  & $.120_{\pm .01}$  & $.487_{\pm .00}$  & \\
$\textsc{{TANL}}$$_{e2e}$ & $.473_{\pm .01}$  & $.343_{\pm .02}$  & $.394_{\pm .00}$  & \\
$\textsc{{DEGREE}}$$_{e2e}$ & $.514_{\pm .01}$  & $.270_{\pm .02}$ & $.595_{\pm .01}$  & \\
$\textsc{{GTT}}$$_{e2e}$ & $.304_{\pm .01}$  & \multicolumn{1}{c}{---}  & $.310_{\pm .00}$  & \\

            \bottomrule
        \end{tabular}
        }
        \caption{\textbf{Event extraction experiments.} {\small 
        Average F1-micro scores on 5-fold cross validation for the prompting baselines, end-to-end and pipeline systems on \muc, \wiki and \cmnee. 
        The colored cells indicate the best performing model, for a given dataset, with the darkest color indicating the highest F1 score. 
        }}
        \label{table:results}
    \end{table*}
    
\section{Models}\label{sec:models}
We study the performance of four state-of-the-art sequence-to-sequence style systems for document-level event extraction: \tanlsys \cite{tanl}, \gttsys \cite{du-etal-2021-template}, \degreesys \cite{hsu-etal-2022-degree}, and \uiucsys \cite{li-etal-2021-document} and compare them against prompt-based baselines using \textsc{GPT-4o} and \textsc{GPT-4o-mini}.

\subsection{Models Overview}
\paragraph{\textsc{GPT-4o} Baselines}
We prompt \textsc{GPT-4o} and \textsc{GPT-4o-mini} to extract event templates in a JSON format giving it six in-context examples. 
We extract all events for all event types using a single query. We use event descriptions and templates as instructions. An example of the prompt can be found in Appendix~\ref{appendix:promting}.

\vspace*{-0.05in}
\paragraph{\tanlsys}\hspace*{-3mm}\cite{tanl} 
is a pipeline-based model that takes only the document as input. 
As shown in Figure~\ref{fig:models-arch}, \tanlsys fine-tunes a \textsc{T5}-large transformer \cite{10.5555/3455716.3455856} to complete event extraction in two phases: 1) event detection, and 2) argument extraction. First, the document is passed through \textsc{T5} to tag/delimit a trigger for each detected event (the entire document is produced as output with event type tags associated with recognized triggers). Then, each identified \{trigger, event type\} pair (tagged/delimited in the input document) is passed through T5 once to identify all event arguments and assign them a role type. 

\vspace*{-0.05in}
\paragraph{\textsc{GTT}}\hspace*{-3mm}\cite{du-etal-2021-template} 
is an end-to-end model that takes a document and an ordered list of the possible event types and generates filled templates for all event types in a single pass through BERT \cite{devlin-etal-2019-bert}. 
Templates are linearized to sequences with special argument and event delimiters.

\vspace*{-0.05in}
\paragraph{\textsc{Degree}}\hspace*{-3mm}\cite{hsu-etal-2022-degree} 
is an end-to-end model that takes a document, a natural language template for one event type \textit{E} and extra information about events of type \textit{E} (i.e.\ type definitions, example triggers); it generates filled natural language templates (that include identified triggers) for all detected events for type \textit{E} in a single pass. The model is called once for each event type of interest for each document. For this reason, training is augmented with multiple negative examples (i.e.\ document-(empty)template pairs for event types that do not appear in the text).

\vspace*{-0.07in}
\paragraph{\textsc{GenIE}} \cite{li-etal-2021-document} 
was originally proposed for argument extraction. Specifically, a document is provided with the trigger delimited by special tokens and a natural language template (similar to that of \degreesys) corresponding to the event type to extract arguments for. The model output is the natural language template filled with arguments. We use \uiucsys for the full event extraction task by using \tanlsys as the trigger extractor and \uiucsys as the argument extractor.


\subsection{Model Type: Pipeline vs End-to-End (e2e), No Trigger}\label{sec:model-type}
We explore both pipeline and end-to-end model architectures and modify \tanlsys and \degreesys to run in both settings.
As shown in Figure~\ref{fig:models-arch}, a pipeline system consists of a trigger extraction and an argument extraction component. The trigger extraction phase identifies all event triggers and their corresponding event types. The argument extraction phase then operates on each \{trigger, event type\} pair separately. This design allows argument extraction to anchor in an informative part of the text, i.e.\ in the vicinity of a known trigger.
End-to-End systems, on the other hand, jointly extract triggers, event types, and arguments, allowing the two extraction tasks to inform one another, thus alleviating cascading errors that can occur in a pipelined setting. 

We next describe changes made to \gttsys, \degreesys and \tanlsys to run in various settings. 

\vspace*{-0.03in}
\paragraph{Pipeline} We modify \degreesys to fit the pipeline architecture.
We train distinct models for trigger and argument extraction. Specifically, the trigger extraction input and output templates only contain slots for a trigger and event type; the argument extraction input templates have an additional slot for a trigger (identified during trigger extraction) and the output templates only remove the slot for the trigger, retaining only slots for each argument type. The model is called once to fill natural language templates for all identified triggers (trigger extraction) and once for each \{identified trigger, event type\} pair to identify arguments (argument extraction). We did not use \gttsys for pipelined experiments. By default, \tanlsys, \tanlsys + \uiucsys are pipelined models.

\paragraph{No Triggers} We modify all models to run without the use of triggers. All \textit{no trigger} experiments were conducted in an e2e manner.
For \tanlsys, the raw text is provided and the model tags arguments with an event type and number and the argument role (e.g.\ ``perpetrating individual for attack 1''). A single span can have multiple such tags if it is a shared entity mention for multiple events. The spans associated with each unique \{event type, number\} pair are processed into fillers for single events. This scheme only requires one pass of T5 for full event extraction. For \degreesys, we only remove the slot for the trigger in input/output templates. By default, \gttsys is an e2e system that does not use triggers.

\paragraph{End-to-End (with triggers)} For \tanlsys, the event type and number and the argument role tag is supplemented with a tag for the trigger (e.g.\ ``perpetrating individual for attack 1 | trigger = killed''). There is still only one pass for full event extraction. For \gttsys, a slot is added at the start of the output event templates for a trigger. By default, \degreesys is an e2e system using triggers.

\section{Experimental Setup}\label{sec:experiments}

In our study of the utility of triggers on document-level event extraction, we tested the performance of fine-tuned (i.e.\ pipeline and end-to-end) and prompt-based approaches on the three datasets described in Section~\ref{sec:dataset}. We consider the prompt-based approaches as a baseline and compare them to fine-tuning the four state-of-the-art models, described in Section~\ref{sec:models}; each model was tested in a pipeline and an end-to-end manner as applicable. 
We train on 1 $\times$ A100 and 4 $\times$ 1080Ti using the default learning rate suggested for each model using the first 10\% of the steps as warmup. The reported score on the test data is obtained using the system from the best performing epoch using the validation set. 


Due to the varying sizes of the datasets, with some very small, we perform 5-cross validation on the test data by sampling 80\% of the data for each evaluation run. 
Each model is trained once and evaluated once on each of the 5 validation splits. We then report average and standard deviation scores across the 5 splits.
Our evaluation focuses on Micro-F1 --- combined F1 score on both event type detection and argument role extraction. 
Since the datasets are multi-event, we have to align the predicted and gold templates.
We follow the matching algorithm of \citet{das-etal-2022-automatic} to align and score the predicted templates\footnote{We modify their proposed scorer by improving its compute time (using the Hungarian Matching algorithm \cite{hungarian}) and by removing span-based scoring as not all models predict spans.}.




\section{Results and Discussion}\label{sec:result-discussion}

In this section, we discuss our experimental results shown in Table~\ref{table:results} (follows Conclusions). We first analyze some high level performance
trends independent of trigger usage before commenting on how triggers should be used depending
on various system and corpora specifications.

\subsection{General Observations for Document-Level Event Extraction}


\paragraph{Result \#1 -- Prompt-based in-context learning approaches for LLMs do not outperform smaller-scale fine-tuned models.}
As shown in Table~\ref{table:results}, 
the best prompt-based systems 
achieve 43.8\% (with human triggers), 38.3\% (keyword), and 66.7\% (human) F1 while the best fine-tuning methods achieve 59.7\% ($\textsc{{DEGREE}}_{e2e}$ no-trigger), 47.1\% ($\textsc{{TANL}}_{pipeline}$ human), and 78.9\% F1 ($\textsc{{DEGREE}}_{e2e}$ no trigger) 
on \muc, \wiki and \cmnee, respectively.

\begin{figure}[tbh!] 
    \includegraphics[width=\columnwidth]{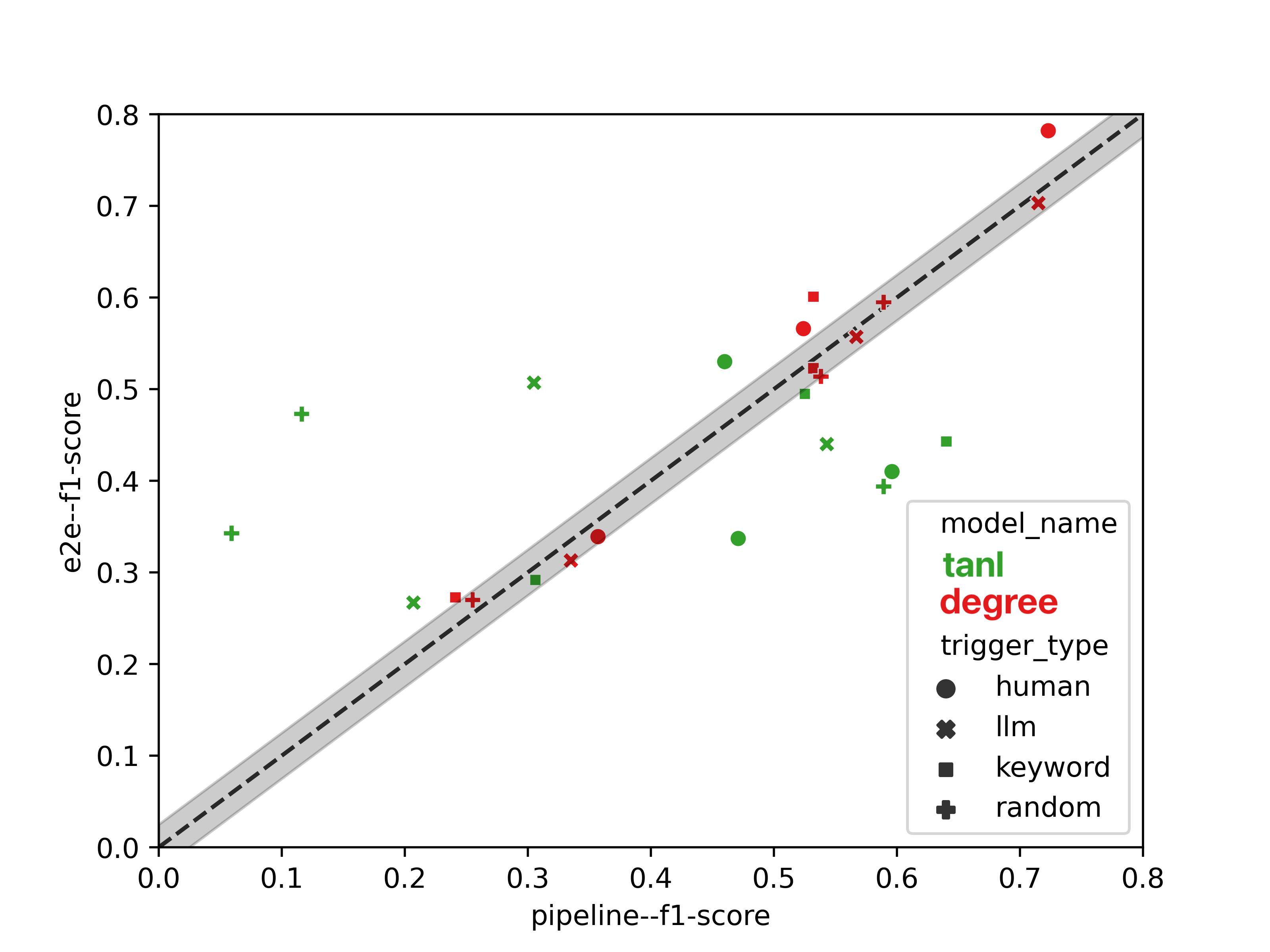}
    \centering
    \small
    \caption{
    \small
    \textbf{E2E vs. Pipeline.} F1-score of each e2e experiment vs.\ its pipeline version. If the data point is above $y=x$ (dashed line), then the e2e architecture outperformed the pipeline architecture for that given (model, trigger type, dataset) setting. (Datasets cannot be distinguished in the figure.)
    }
    \label{fig:e2e-vs-pipeline}
\end{figure}

\paragraph{Result \#2 -- End-to-end systems perform similarly to pipeline architectures for the same model types.}

As shown in Figure~\ref{fig:e2e-vs-pipeline}, 
e2e models trained with triggers perform roughly on par or better than their pipelined counterparts with five explainable exceptions. 
All exceptions involve the \tanlsys system applied to \wiki and \cmnee. We hypothesize that since event arguments are localized around the trigger for both datasets (see Table~\ref{table:datastat-stats}), dedicating a separate pipeline component towards trigger extraction, which helps disambiguate events and anchor argument extraction, is particularly beneficial. (This is explored in more detail in Section \ref{sec:claim_3} where the propensity of \degreesys to ignore triggers is also discussed.)
%
%
This is important because we can generally build e2e systems that are faster to train and quicker during inference than pipelined systems without sacrificing performance.

\subsection{Observations on the Importance of \triggerplural~ for Document-Level Event Extraction}\label{sec:discussion-triggers}

\paragraph{Result \#3 -- The presence of triggers, even random ones, is generally \textit{needed} for prompt-based approaches.}

Across both \textsc{GPT} models and multiple datasets, we observe that in-context prompting with triggers (from \textit{any} trigger source) regularly outperforms the associated no-trigger variations. This is especially drastic for \textsc{GPT-4o}  on \wiki, where we see a 5 point F1 gain after incorporating keyword triggers in the in-context examples as shown in Table~\ref{table:results}. 
Interestingly, even adding random spans as triggers boosts performance by ~2 points for \textsc{GPT-4o}.
Upon manual inspection
we observe that \textsc{GPT-4o} actually identifies legitimate trigger-like phrases in spite of the nonsense triggers in the in-context examples.
This follows findings in recent work \cite{min2022rethinkingroledemonstrationsmakes} and suggests that learning the concept of a trigger, even without proper examples, meaningfully supports extraction. 

\begin{figure*}[hbt!] 
    \centering
    \small
    \begin{subfigure}[t]{0.49\textwidth}
        \centering
        \includegraphics[width=\columnwidth]{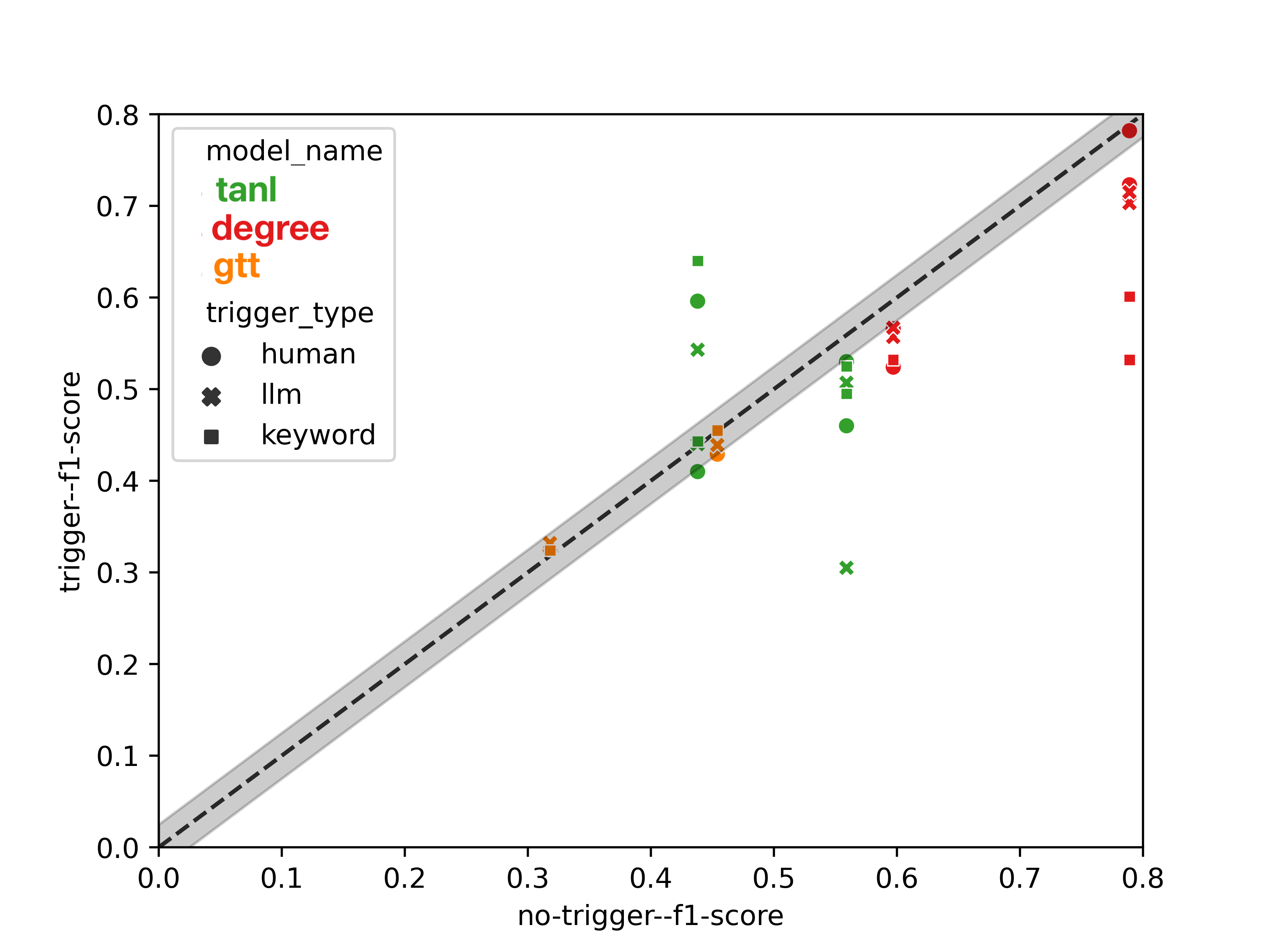}
        \caption{
        \small \muc \& \cmnee
        }
        \label{fig:all-vs-no-on-cmnee-muc}
    \end{subfigure}
    ~ 
    \begin{subfigure}[t]{0.49\textwidth}
        \centering
        \includegraphics[width=\columnwidth]{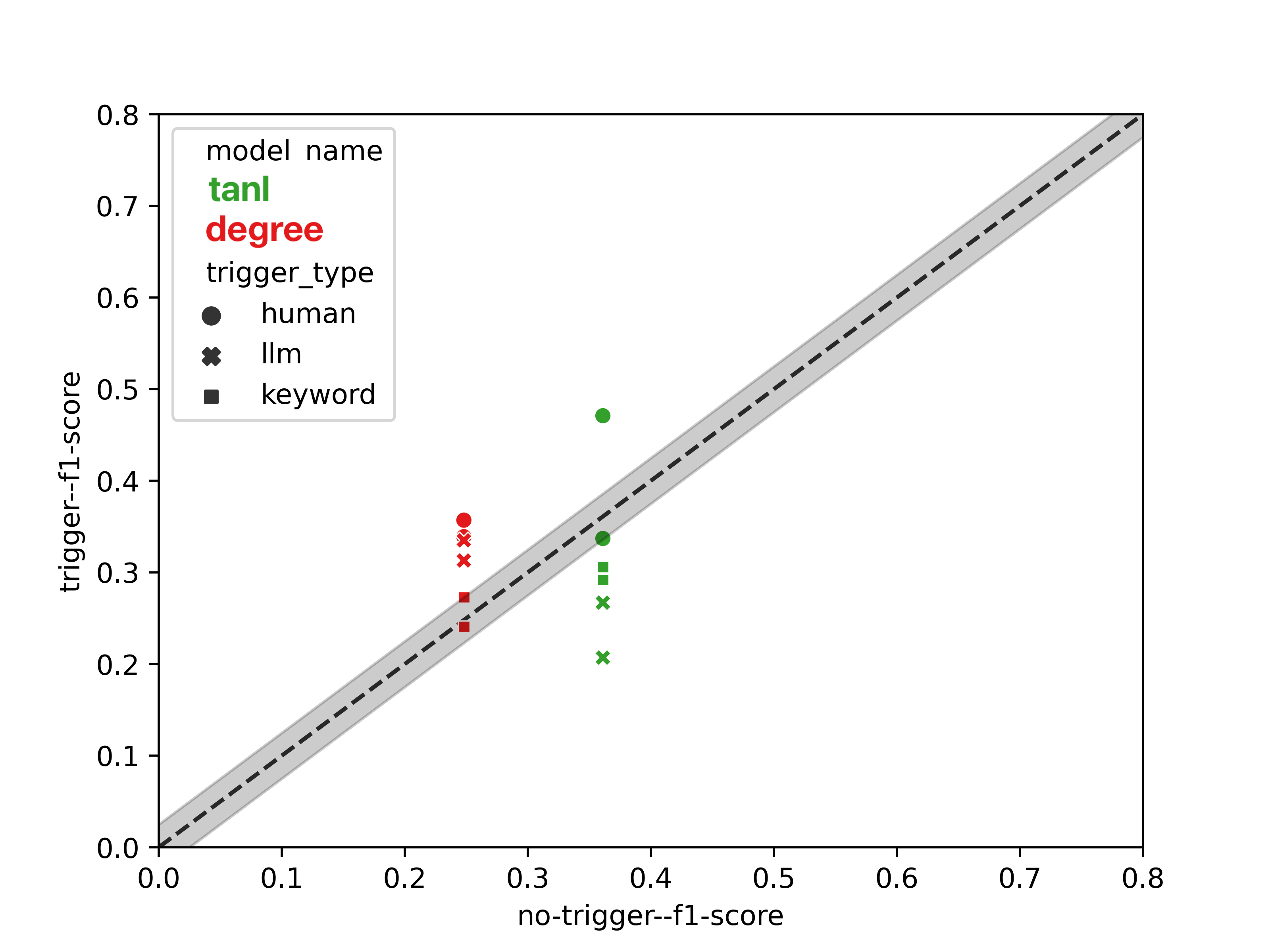}
        \caption{\small \wiki}
        \label{fig:all-vs-no-on-wiki}
    \end{subfigure}
    \caption{
    \small
    \textbf{Triggers vs. no-Triggers.} F1-score of each fine-tuning experiment with triggers (annotated by humans, LLMs, or the keyword heuristic) against its e2e version without triggers. A data point above $y=x$ (dashed line) implies that fine-tuning with triggers performed better for that (model, architecture, dataset).
    }
\end{figure*}

\paragraph{Result \#4 -- Triggers are \textit{not} needed for corpora that have few events to extract in each document.}\label{par:claim_2}



We notice for \cmnee and (especially) \muc that systems trained without triggers perform comparably to systems trained with triggers (Figure \ref{fig:all-vs-no-on-cmnee-muc}) (i.e.\ most points fall below the $x=y$ diagonal). The key commonality between the two datasets is their relatively low number of events per document (compared to \wiki). We hypothesize that event disambiguation is therefore easier in \muc and \cmnee; therefore the use of triggers (to anchor argument extraction) for these two datasets is less useful. Exceptions to this occur when applying \tanlsys to \cmnee where fine-tuning in the pipelined setting performs better with triggers (e.g., 54.3\%, 59.6\% and 64.0\% F1) than without (43.8\% F1). We believe this may be due to localized events for \cmnee as noted in Result \#2 and Result \#5.

\paragraph{Result \#5 -- Higher quality triggers are preferred for corpora that have many \textit{localized} events to extract in each document.}\label{sec:claim_3}

This result is almost a negation 
of the previous claim, except for the additional conditions of \textit{localized} events and \textit{higher quality} triggers. Aside from the typical number of events per document, \wiki also differs from \muc and \cmnee in how closely grouped together event arguments are (see Table~\ref{table:datastat-stats}). We unfortunately do not have a dataset with documents with many events with dispersed arguments to explore the many event per document setting more generally. In Figure~\ref{fig:all-vs-no-on-wiki} and Table~\ref{table:results}, we see that e2e and pipelined systems for \wiki fine-tuned with human triggers perform better than their fine-tuned counterparts without triggers. Our conclusion regarding the number of events per document is similar to Result \#2: many events require careful event disambiguation and identifying relevant triggers serves this purpose well. Regarding event localization, the trigger's position is a strong clue to inform systems where to look for associated arguments. In fact, we see that pipelined \tanlsys and \tanlsys + \uiucsys, the only systems that have access to the exact location of triggers within a document prior to argument extraction, are the best models for \wiki (see Table\ref{table:results}).

\begin{figure*}[hbt!]
    \centering
    \small
    \begin{subfigure}[t]{0.49\textwidth}
        \centering
        \includegraphics[width=\columnwidth]{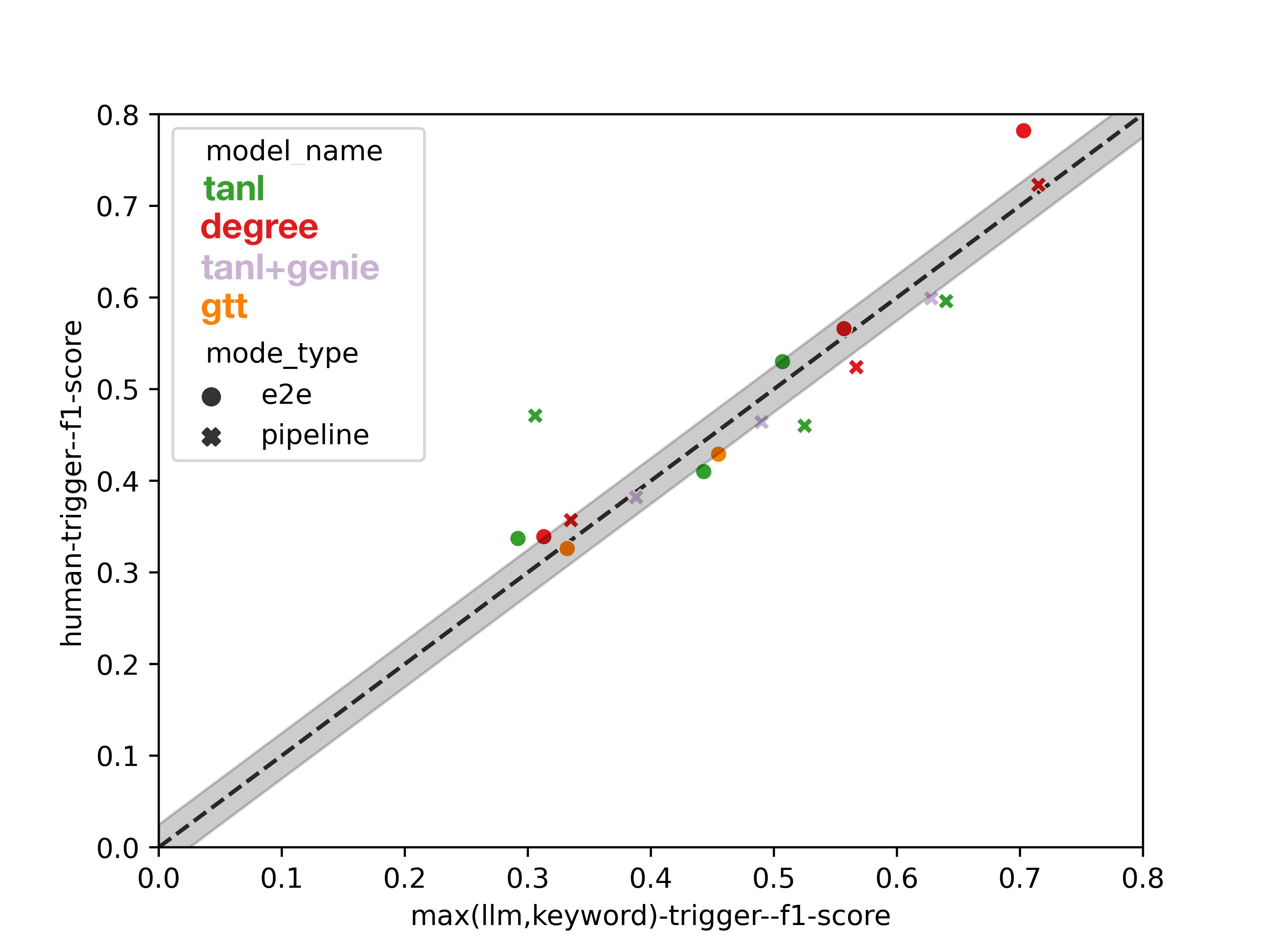}
        \caption{
        \small
        Against/ \textsc{Human}
        }
        \label{fig:human-vs-all-alldatasets}
    \end{subfigure}
    ~ 
    \begin{subfigure}[t]{0.49\textwidth}
        \centering
        \includegraphics[width=\columnwidth]{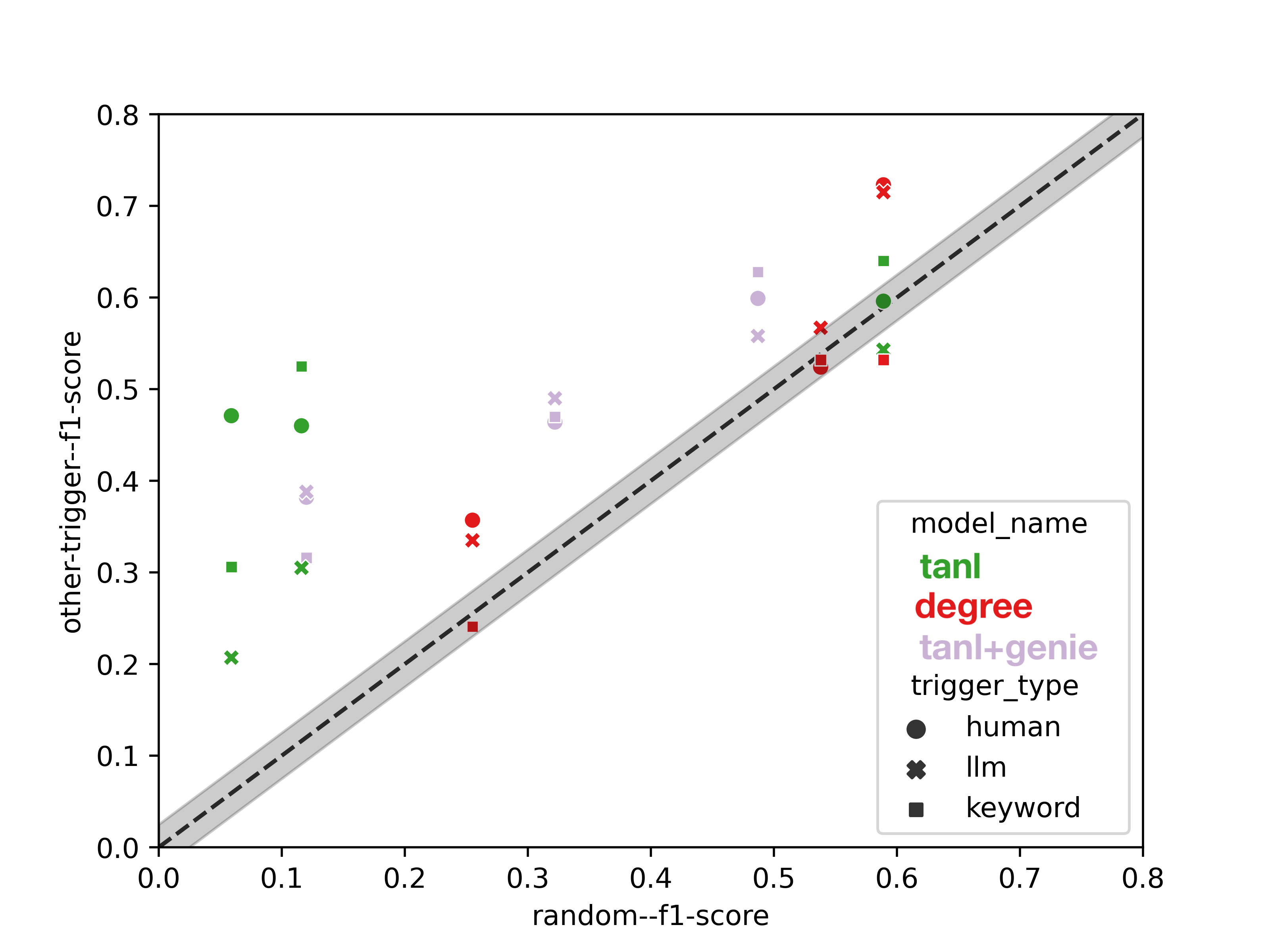}
        \caption{
        \small
        Against/ \textsc{Random}
        }
        \label{fig:random-vs-other}
    \end{subfigure}
    \caption{
    \small
    \textbf{Trigger vs. Trigger.} This plots the F1-scores of experiments fine-tuning with one trigger source against another. In \ref{fig:human-vs-all-alldatasets}, a point above $y=x$ (dashed line) implies that using human annotations performed better than using any other annotation source, while in \ref{fig:random-vs-other} and using either human, LLM, or keyword annotations performed better than random annotations.
    }
\end{figure*}

\paragraph{Result \#6 -- Lower quality trigger annotations that are
less expensive to obtain \textit{can} be used as an alternative to 
\textsc{human} triggers.}


We see in Figure~\ref{fig:human-vs-all-alldatasets} that trigger quality does not significantly impact overall performance for fine-tuning methods. This suggests that models can learn textual cues beyond triggers during extraction. An exception to this is the pipelined \tanlsys system applied to \wiki, though we believe this relates to the discussion for Result \#5.

\paragraph{Result \#7 -- Robustness with respect to degraded trigger quality can be achieved by giving the event extraction system access to additional event information (e.g. templates, instructions).}  
Evidence for this can be seen by examining the performance of \degreesys, which does not exhibit any significant swings in overall performance when comparing across trigger annotation sources. For \muc and \cmnee, \degreesys actually performs best (and outperforms all other systems) in the \textit{no triggers} setting (59.7\% and 78.9\% F1, respectively). We believe
that this robustness arises from the background information it requires as part of its input, including a natural language statement of the template for the target event type, a definition of the event type, a short list of likely triggers, short definitions of expected event arguments. 
As additional evidence for this claim, one can observe the \textit{random triggers} results for \textsc{{TANL}}$_{pipeline}$ in Table~\ref{table:results}, which are significantly
worse than its e2e variation (11.6\% versus 47.3\%,  5.9\% versus 34.3\%, 22.3\% versus 39.4\% F1 for \muc, \wiki, and \cmnee, respectively). This difference, however, is remedied by providing a natural language template as input for argument extraction, as in the case of \tanlsys + \uiucsys: the addition results in +20.6, +6.1, and +26.4 F1 \% points for \muc, \wiki, and \cmnee, respectively. 
These results suggest that additional relevant inputs such as natural language templates that relate argument types and descriptions of events benefit fine-tuning methods to the extent of possibly overcoming systematic flaws such as conditioning extraction on noise. Overall, we see the trend outlined in Figure~\ref{fig:random-vs-other}: \tanlsys exhibits very poor performance, adding event template information (\tanlsys + \uiucsys) improves system performance, and the inclusion of event-specific instructions (\degreesys) allows  further robustness when training on random triggers.

\section{Limitations}\label{sec:limitations}
Although many current state of the art approaches employ some sequence-to-sequence language modeling approach, there are alternative models that we did not investigate. For example, \textsc{IterX} \cite{chen-etal-2023-iterative} formulates event extraction as a Markov decision process while \textsc{Procnet} builds a graph over entity representations to be parsed into events \cite{wang-etal-2023-document}. Additionally, some more elaborate prompting-based approaches such as \textsc{Code4Struct} \cite{wang2022code4struct} warrant testing given the rapidly developing capabilities of LLMs.

\section{Conclusions and Future Work}\label{sec:conclusion}
In this work, we explore the use of triggers in document-level event extraction. Although typically regarded as an integral part of sentence-level event extraction, we showed human annotated triggers are only necessary given certain dataset and modeling parameters (many localized events in a document, providing relevant conditioning information such as natural language templates). 

Future work could investigate a wider scope of models, as discussed just above in Section~\ref{sec:limitations} and/or incorporate additional datasets where available. Existing datasets such as \cmnee and \textit{DCFEE} \cite{yang-etal-2018-dcfee} are in Chinese (therefore likely require translation before use) while others such as \textit{RAMS} \cite{ebner-etal-2020-multi} and \textit{Doc-EE} \cite{tong-etal-2022-docee} do not exhibit traits of document level datasets (\textit{RAMS} has 1 event per document and ``document" are restricted to 5 sentence windows surrounding the trigger while \textit{Doc-EE} has only one event per document). Efforts creating new document level event extraction datasets would significantly expand work like ours that can inform future event extraction system design choices.


\onecolumn

\bibliography{tacl2021}
\bibliographystyle{acl_natbib}

\newpage
\appendix
\section{Annotation Guidelines}
\label{appendix:annotation-guideline}

Similar to \citealp{ace05}, we define a trigger as a word that indicate the occurrence of an event. 
A trigger can be any part of speech --- e.g.\ verb, noun or adjective; but the majority are verbs and nouns. Common triggers for the different event-types in the \muc dataset include ``attack", ``bomb", ``blew up", ``killed", etc.  
In Table~\ref{table:annotation-examples}, we show the additional annotation guidelines we curated to annotate trigger-span(s) for \muc. 

\begin{table}[tbhp]
    
    \centering
    \resizebox{0.8\columnwidth}{!}{
    \begin{tabular}{cp{0.5\linewidth}p{0.35\linewidth}}
      \toprule
       & \multicolumn{1}{c}{\bf Rule Description} & \multicolumn{1}{c}{\bf Example} \\
      \midrule
      1 & \textit{Keep trigger spans as long as possible, but divide spans if uninformative words fall between keywords} & \underline{dynamite charge}$_{\text{trigger}1}$ today \underline{exploded}$_{\text{trigger}2}$ $\cdots$ \\ \\
      \hline
      2 & \textit{Do not include statistics in triggers} & \underline{attack}$_{\text{trigger}1}$ in which 20 persons were \underline{killed}$_{\text{trigger}2}$ and approximately 100 were \underline{injured}$_{\text{trigger}3}$ $\cdots$ \\ \\
      \hline
      3 & \textit{Do no include adjectives in triggers} & \underline{threatening}$_{\text{trigger}1}$ them with hand \underline{grenades}$_{\text{trigger}2}$ $\cdots$ \\ \\
      \hline
      4 & \textit{Do not include recipients/direct objects in triggers} &  \underline{murder}$_{\text{trigger}1}$ of Jecar Neghme $\cdots$ \\ \\
      \hline
      5 & \textit{Do not annotate words such as “body”, “bodies”, “injury”, “injured”, “wounded” unless there are no other triggers present in the paragraph} & \\ 
      \hline
      6 & \textit{Do not include words that indicate verb tense such as “will”, “is”, “are”, “have”} & \\
      \bottomrule
    \end{tabular}
    }
    \caption{\textbf{Annotation Guidelines} we describe the additional rules and definitions we used to trigger in conjecture with \citealp{ace05}. We give example to each rule and \underline{indicate} which tokens are selected as triggers.}
    \label{table:annotation-examples}
  \end{table}
\section{Prompting}
\label{appendix:promting}

In this section we show the prompts there were used for our experiments. 
In the following table, we show the prompt that was used to perform document-level event extraction for \muc. For different trigger sources, we would change the trigger used and remove the entry completely when running the no \triggers experiment. \\

\begin{lstlisting}
----------- System Prompt ----------
You are journalist reading news articles and identifying the main events that the article is reporting...

-------- In Context Examples -------
Perform event extraction on the provided document and identify key details (such as event-type, involved organizations or individuals, target of the event, victim of the event, and weapon used) and output it in a json format.

Document:
   san salvador, 12 nov 89 (domestic service) -- [communique] [salvadoran armed forces press committee] [text] the salvadoran armed forces inform the people of el salvador that fmln [farabundo marti national liberation front] ...  

Events:
```[
  { 
    "trigger": "vandals",
    "event type": "attack",
    "perpetrator individual": ["groups of vandals"],
    "perpetrator organization": ["fmln"],
    "target": ["shopping centers"],
    "victim": [],
    "weapon": []
  }
]```

...

Perform event extraction on the provided document ...

Document:
   lima, 10 jan 90 (efe) -- [text] the national police reported today that over 15,000 people have been arrested in lima over the past few hours ...

Events:

----------- GPT response -----------
```[
  { 
    "trigger": "was riddled with bullets",
    "event_type": "attack",
    "perpetrating individual": ["gerardo olivos silva"],
    "perpetrating organization": ["shining path"],
    "target": ["enrique lopez albujar trint"],
    "victim": ["enrique lopez albujar trint"],
    "weapon": ["firearm"]
  }
  ...
]```
\end{lstlisting}

\iftaclpubformat
\fi

\end{document}